\documentclass[10pt,twocolumn,letterpaper]{article}

\usepackage{iccv}            

\usepackage{wrapfig}
\usepackage{url}
\usepackage{booktabs}
\usepackage{amsfonts}
\usepackage{nicefrac}
\usepackage{microtype}
\usepackage{lipsum}
\usepackage{fancyhdr} 
\usepackage{graphicx}
\usepackage{appendix}
\usepackage{algorithmic}
\usepackage{color,soul}
\sethlcolor{yellow}
\usepackage{multirow}
\usepackage{amsmath}
\usepackage{listings}
\usepackage{verbatim}
\usepackage{float}
\usepackage{subcaption}
\usepackage{comment}
\usepackage{dsfont}
\usepackage{algorithm}
\usepackage{booktabs}
\usepackage{caption} 
\usepackage{xcolor}
\usepackage{colortbl} 
\usepackage{threeparttable}
\definecolor{darkgreen}{RGB}{0,190,0}
\usepackage{amssymb}
\usepackage{pifont}
\usepackage{tikz}
\usetikzlibrary{decorations.pathreplacing, positioning}
\definecolor{mygreen}{RGB}{130,179,102}
\definecolor{myblue}{RGB}{108,142,191}
\definecolor{cvprorange}{RGB}{217,95,2} 

\usepackage{etoolbox}
\makeatletter
\AfterEndEnvironment{algorithm}{\let\@algcomment\relax}
\AtEndEnvironment{algorithm}{\kern2pt\hrule\relax\vskip3pt\@algcomment}
\let\@algcomment\relax
\newcommand\algcomment[1]{\def\@algcomment{\footnotesize#1}}
\renewcommand\fs@ruled{\def\@fs@cfont{\bfseries}\let\@fs@capt\floatc@ruled
  \def\@fs@pre{\hrule height.8pt depth0pt \kern2pt}%
  \def\@fs@post{}%
  \def\@fs@mid{\kern2pt\hrule\kern2pt}%
  \let\@fs@iftopcapt\iftrue}
\makeatother

\definecolor{iccvblue}{rgb}{0.21,0.49,0.74}
\usepackage[pagebackref,breaklinks,colorlinks,allcolors=iccvblue]{hyperref}
\usepackage{color}

\def\paperID{10} 
\def\confName{ICCV}
\def\confYear{2025}

\title{
Fake \& Square: Training Self-Supervised Vision Transformers \\ with Synthetic Data and Synthetic Hard Negatives
}

\author{Nikolaos Giakoumoglou \quad Andreas Floros \quad Kleanthis Marios Papadopoulos \quad Tania Stathaki\\
Imperial College London\\
{\tt\small \{nikos, andreas.floros18, kleanthis-marios.papadopoulos18, t.stathaki\}@imperial.ac.uk}\\
}

\begin{document}
\maketitle


\begin{abstract}

This paper does not introduce a new method per se. Instead, we build on existing self-supervised learning approaches for vision, drawing inspiration from the adage "fake it till you make it". While contrastive self-supervised learning has achieved remarkable success, it typically relies on vast amounts of real-world data and carefully curated hard negatives. To explore alternatives to these requirements, we investigate two forms of "faking it" in vision transformers. First, we study the potential of generative models for unsupervised representation learning, leveraging synthetic data to augment sample diversity. Second, we examine the feasibility of generating synthetic hard negatives in the representation space, creating diverse and challenging contrasts. Our framework—dubbed $\text{Syn}^2\text{Co}$—combines both approaches and evaluates whether synthetically enhanced training can lead to more robust and transferable visual representations on DeiT-S and Swin-T architectures. Our findings highlight the promise and limitations of synthetic data in self-supervised learning, offering insights for future work in this direction.

\end{abstract}


\section{Introduction}
\label{sec:introduction}

The field of computer vision has witnessed remarkable progress in recent years. The progress of AI systems has been fueled by large-scale datasets that are carefully annotated by humans. However, the reliance on manually labeled data has become a bottleneck for further advancements, leading to increased interest in self-supervised learning methods that leverage vast amounts of unlabeled data \cite{liu2021self}. Deep learning models are particularly data-hungry, requiring vast amounts of labeled data to perform well, yet the rapid expansion of available data has outpaced our capacity to label it comprehensively. Contrastive methods \cite{chen2020simclr,he2020moco} have demonstrated that self-supervised learning can rival and even surpass supervised pre-training, particularly in tasks requiring robust feature representations.

\begin{figure}[t]
    \centering
    \includegraphics[width=0.478\textwidth]{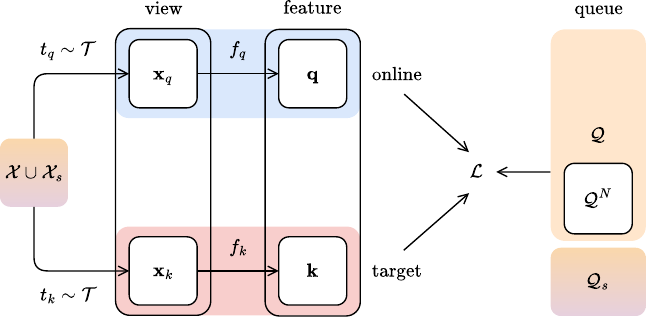}
    \caption{Overview of our pipeline for contrastive learning. It includes synthetic data, $\mathcal{X}_s$, produced by a diffusion model and synthetic negatives, $\mathcal{Q}_s$, which are generated on the fly.}
    \label{fig:overview}
\end{figure}

\begin{figure*}[!t]
    \centering
    \begin{minipage}[t]{0.49\textwidth}
        \centering
        \raisebox{0pt}[\height][\depth]{\includegraphics[width=\textwidth]{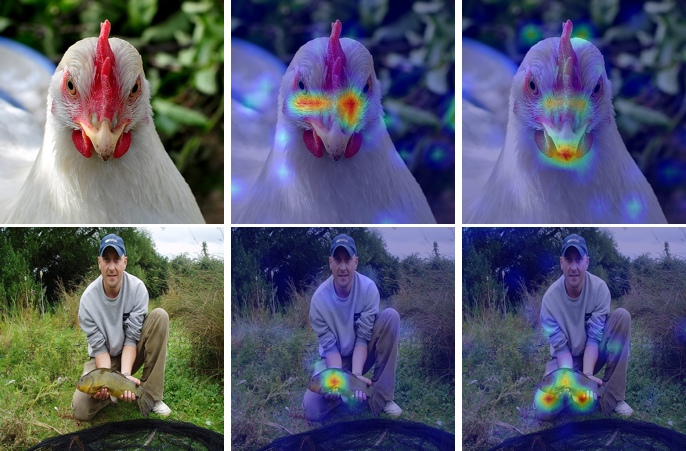}}
        \caption{DeiT Grad-CAMs. Left: input image. Middle: pretraining with real data. Right: pretraining with synthetic data.}
        \label{fig:segm}
    \end{minipage}
    \hfill
    \begin{minipage}[t]{0.49\textwidth}
        \centering
        \raisebox{0pt}[\height][\depth]{\includegraphics[width=\textwidth]{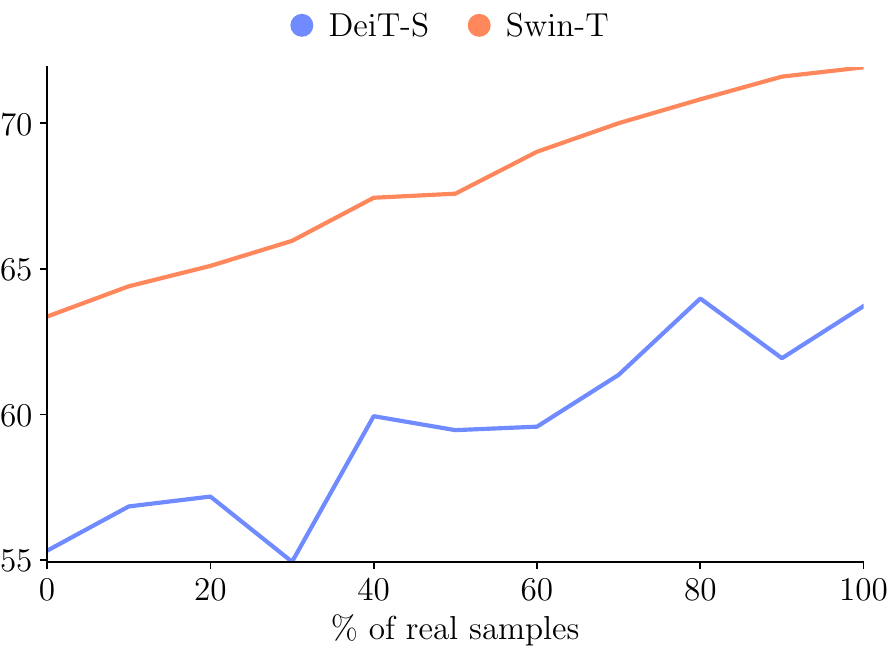}}
        \caption{Top-1 linear probing accuracy on ImageNet-100. A percentage of the pretraining data is real and the rest is synthetic.}
        \label{fig:effect_synthetic_data}
    \end{minipage}
\end{figure*}

Despite their success, contrastive learning methods face two significant challenges. First, they often struggle with the limited diversity of hard negative examples, which are crucial for learning discriminative features \cite{kalantidis2020mochi,zhang2021understandinghardnegativesnoise}. Second, they typically require vast amounts of real-world data to achieve competitive performance, creating a data bottleneck similar to supervised learning, limiting scalability, generalization, and applicability to low-resource domains \cite{giakoumoglou2024review}.

We are inspired by the adage: \textit{"fake it till you make it"} – presenting something "\textit{synthetic}" as if it were "\textit{real}" until it actually produces results. We apply this concept to unsupervised contrastive learning in two ways: by introducing synthetic data clones in the image space, effectively expanding the training dataset, and by generating synthetic hard negatives in the representation space, creating challenging contrasts. We focus on vision transformers, which have achieved remarkable success, surpassing ResNets \cite{chen2022visiontransformersoutperformresnets}.

The main \textbf{contributions} of our work are as follows:

\begin{itemize}
    \item We create a synthetic clone of ImageNet-100 and demonstrate how synthetic data affects contrastive self-supervised learning.
    \item We adapt and extend synthetic negative generation to transformers, showing that DeiT and Swin can leverage synthetic contrasts to learn more discriminative features.
\end{itemize}


\section{Related Work}\label{sec:related}

\paragraph{Self-supervised visual representation learning.} 

Self-supervised learning has improved visual representation learning through various paradigms \cite{giakoumoglou2024review}. Recent methods include contrastive learning ~\cite{chen2020simclr,he2020moco}, clustering~\cite{caron2020swav,caron2019deepcluster}, and distillation approaches~\cite{grill2020byol,caron2021dino}. While SimCLR~\cite{chen2020simclr} established the importance of augmentation strategies and large batch sizes, MoCo~\cite{he2020moco} introduced memory banks for scalable negative sampling.  The quality of negative samples in contrastive learning has been a focus of extensive research \cite{arora2019theoretical,kalantidis2020mochi,giakoumoglou2024synco}. These studies aim to select informative negative samples and address false negatives in instance discrimination tasks.

\paragraph{Transformers for vision.} 

The success of Transformers \cite{vaswani2017attention} in NLP has inspired continuous efforts to generalize this architecture to computer vision tasks \cite{dosovitskiy2021vit,touvron2021deit,liu2021swin}. ViT \cite{dosovitskiy2021vit} demonstrated that a pure transformer applied to sequences of image patches can excel in image classification tasks, particularly when pre-trained on large datasets. DeiT \cite{touvron2021deit} further showed that ViTs can be effectively trained on ImageNet without requiring external data. Swin Transformer \cite{liu2021swin} introduced a hierarchical design with shifted windows, making it more suitable for dense prediction tasks. Self-supervised learning for vision transformers has rapidly evolved, including contrastive adaptations \cite{chen2021mocov3,xie2021moby}, masked image modeling methods \cite{he2021mae}, and distillation frameworks \cite{caron2021dino,zhou2022ibot}.

\paragraph{Synthetic data for vision.} 

Synthetic data generation has become increasingly important in computer vision \cite{nikolenko2021synthetic}. This field has advanced significantly through GANs that produce highly realistic images \cite{karras2019style}. Recent research has explored generating synthetic data specifically for ImageNet classes \cite{deng2009imagenet} using class-conditional GANs \cite{besnier2020dataset}. Beyond generation, self-supervised learning methods have effectively leveraged synthetic data, with SynCLR \cite{tian2023learning} demonstrating that learning from synthetic images can rival results obtained from real data. Similarly, \citet{sariyildiz2023fake} showed that models trained on diffusion-generated ImageNet clones can perform comparably to those trained on real data, exhibiting strong generalization capabilities.


\section{Methodology} 
\label{sec:methodology}

In this section, we present our approach to incorporating synthetic elements into self-supervised learning. We formulate the task (\Cref{sec:task_formulation}), describe the contrastive learning framework (\Cref{sec:contrastive_learning}), and detail our methods for generating synthetic data (\Cref{sec:synthetic_data}) and synthetic hard negatives (\Cref{sec:synthetic_negatives}). We illustrate our complete method, which we refer to as $\text{Syn}^2\text{Co}$, in \Cref{fig:overview}.


\begin{figure*}[!t]
    \centering
    \begin{subfigure}[t]{0.49\textwidth}
        \centering
        \includegraphics[width=\textwidth]{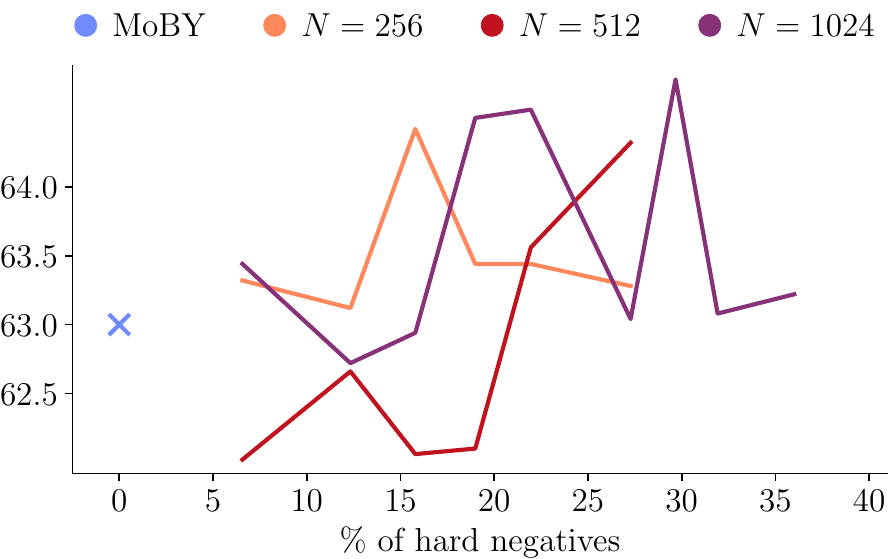}
        \caption{DeiT-S}
         \label{fig:effect_synthetic_negatives_deit}
    \end{subfigure}
    \hfill
    \begin{subfigure}[t]{0.49\textwidth}
        \centering
        \includegraphics[width=\textwidth]{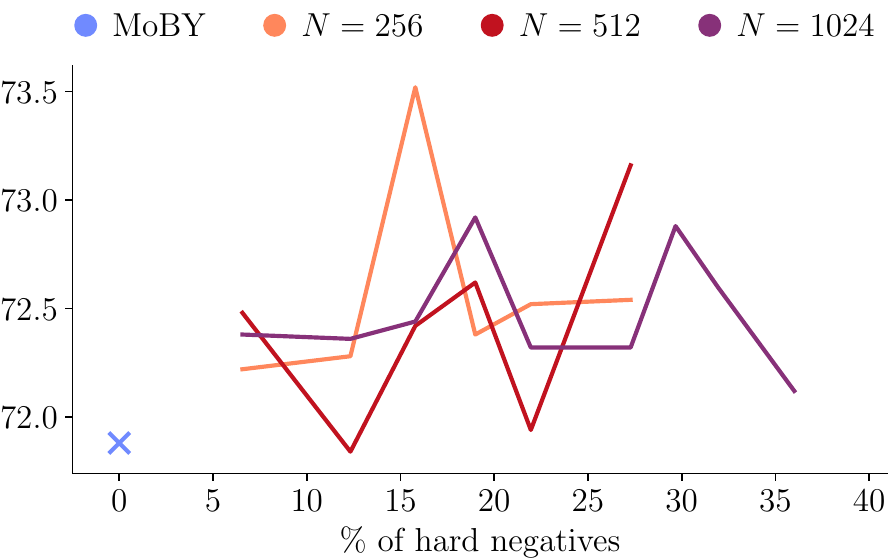}
        \caption{Swin-T}
          \label{fig:effect_synthetic_negatives_swin}
    \end{subfigure}
    \caption{Top-1 linear probing accuracy on ImageNet-100. We pretrain for 100 epochs on real data (\textit{i.e.}, ImageNet-100 training set) using different hardness selection values and synthetic negative percentages.}
    \label{fig:effect_synthetic_negatives}
\end{figure*}

\subsection{Task Formulation}
\label{sec:task_formulation}

Our goal is to learn meaningful representations without relying on labeled data or purely real image datasets. Formally, given a dataset of images, $\mathcal{X}$, we aim to learn a representation function, $f_q:\mathcal{X}\to\mathcal{Z}$, that produces a feature space where semantically similar images are close together and dissimilar images are far apart. 
Unlike standard self-supervised learning, here we explore the possibility of enhancing the training process by augmenting the data with synthetic images, $\mathcal{X}_s$, synthesized using a generative diffusion model $\mathcal{G}$. Additionally, we use synthetic hard negatives, $\mathcal{Q}_s$, constructed in the representation space.

\subsection{Contrastive Learning} 
\label{sec:contrastive_learning}

We employ a contrastive approach to learn representations by comparing similar (\textit{positives}) and dissimilar (\textit{negatives}) samples. Given an image, $\mathbf{x}\sim\mathcal{X}$, and a family of image augmentations, $\mathcal{T}$, we create two views of the same image: $\mathbf{x}_q = t_q(\mathbf{x})$ and $\mathbf{x}_k = t_k(\mathbf{x})$, where $t_q, t_k \sim \mathcal{T}$. 
These views are then mapped using an \textit{online} encoder (our desired representation function), $f_q$, and a \textit{target} encoder, $f_k$, producing feature vectors $\mathbf{q}=f_q(\mathbf{x}_q)$ and $\mathbf{k}=f_k(\mathbf{x}_k)$. $f_q$ is optimized via backpropagation, by minimizing the following InfoNCE \cite{oord2019cpc} contrastive objective:

\begin{equation}\label{eq:contrastive_loss}
    \mathcal{L} = -\log \frac{\exp(\mathbf{q}^\top \cdot \mathbf{k} / \tau )}{\exp(\mathbf{q}^\top \cdot \mathbf{k} / \tau ) + \sum\limits_{\mathbf{n} \in \mathcal{Q}} \exp(\mathbf{q}^\top \cdot \mathbf{n} / \tau)},
\end{equation}

\noindent where $\mathcal{Q} = \{\mathbf{n}_1, \mathbf{n}_2, \ldots, \mathbf{n}_K\}$ is a set of $K$ negative samples and $\tau$ is the temperature that controls sharpness. Negative samples are either drawn from the current batch \cite{chen2020simclr} or from a memory queue \cite{wu2018instdis,he2020moco}. $f_k$ mostly mirrors $f_q$ and is updated via a momentum mechanism at each step:

\begin{equation}
    \theta_k \leftarrow m \cdot \theta_k + (1-m) \cdot \theta_q.
\end{equation}


\begin{table*}[!t]
\centering
\setlength{\tabcolsep}{1mm}
\begin{threeparttable}
\caption{Top-1 and top-5 linear probing accuracy on ImageNet-100 for different methods. We reproduce all methods.}
\label{tab:main_results_table}
\begin{tabular}{lcccccccc}
\toprule
\multirow{2}{*}{Method} & \multirow{2}{*}{\begin{tabular}{c}Synthetic\\Data\end{tabular}} & 
\multirow{2}{*}{\begin{tabular}{c}Synthetic\\Negatives\end{tabular}} & 
\multirow{2}{*}{Epochs} & \multirow{2}{*}{Iterations} & 
\multicolumn{2}{c}{DeiT-S} & \multicolumn{2}{c}{Swin-T} \\
\cmidrule(lr){6-7}\cmidrule(lr){8-9}
 & & & & & Top-1 (\%) & Top-5 (\%) & Top-1 (\%) & Top-5 (\%) \\
\midrule
\textit{Supervised} & - & - & 300 & 76,200  & 80.50 & 94.20 & 86.78 & 97.14  \\
\midrule
DINO \cite{caron2021dino}\textsuperscript{\textdagger} & - & - & 300 & 304,800 & 75.42 & 93.82 & - & -  \\
MoCo-v3 \cite{chen2021mocov3}\textsuperscript{\textdaggerdbl} & \ding{55} & \ding{55} & 400 & 25,600 & 79.41 & 92.20  & - & - \\
\rowcolor{gray!10} MoBY \cite{xie2021moby} & \ding{55} & \ding{55} & 300 & 76,200 & 79.36 & 95.32   & 83.90 & 96.76 \\
\midrule 
\rowcolor{orange!10} & \ding{55} & \checkmark & 300 & 76,200 & 78.96 & 95.06 & \textbf{84.04} & \textbf{96.94}  \\
\rowcolor{orange!10} & \checkmark & \ding{55} & 150 & 76,200  & 79.02 & 94.94 & 83.45 & 96.01 \\
\rowcolor{orange!10} \multirow{-3}{*}{$\text{Syn}^1\text{Co}^*$} & \checkmark & \ding{55} & 300 & 152,400 & 81.86 & 95.80 & 83.68 & 96.84  \\
\midrule 
\rowcolor{blue!10} & \checkmark & \checkmark & 150 & 76,200  & 78.12 & 94.64 & 83.55 & 96.24 \\
\rowcolor{blue!10} \multirow{-2}{*}{$\text{Syn}^2\text{Co}$} & \checkmark & \checkmark & 300 & 152,400  & \textbf{82.12} & \textbf{95.86} & 83.70 & 96.76 \\
\bottomrule
\end{tabular}
\textsuperscript{\textdagger} {\small Trained with batch size 128.} \quad \textsuperscript{\textdaggerdbl} {\small Trained with batch size 2048.} \\
\textsuperscript{*} {\small When including either synthetic data or synthetic negatives (but not both), we refer to this as $\text{Syn}^1\text{Co}$ for fun.}
\end{threeparttable}
\end{table*}

\subsection{Synthetic Data} 
\label{sec:synthetic_data}

We use an image generator, $\mathcal{G}$, to create a synthetic dataset, $\mathcal{X}_s$, that complements the real dataset $\mathcal{X}$. 
Specifically, we leverage the diffusion model from \cite{teng2024relay} to generate clones of ImageNet-100. This approach results in a synthetic dataset of approximately 130,000 images that can be used either as a replacement for or in conjunction with the real dataset. 
During training, images may be sampled from $\mathcal{X} \cup \mathcal{X}_s$. This mixed sampling strategy enables us to investigate various training scenarios, including training exclusively on real data ($\mathcal{X}$ only), training exclusively on synthetic data ($\mathcal{X}_s$ only), and training on a mixture of real and synthetic data with different mixing ratios (see \Cref{fig:segm} and \Cref{fig:effect_synthetic_data}).

\subsection{Synthetic Hard Negatives}
\label{sec:synthetic_negatives}

We generate synthetic hard negatives directly in the representation space \cite{kalantidis2020mochi,giakoumoglou2024synco,yeh2022dcl} to provide challenging examples that help the model learn more discriminative features. This addresses a key limitation in contrastive learning, \textit{i.e.}, the scarcity of informative negative examples that effectively capture semantic boundaries. 
Given a query representation, $\mathbf{q}\in\mathcal{Z}$, and a negative queue, $\mathcal{Q}$, we first select the top $N$ hardest negatives, $\mathcal{Q}^N \subseteq \mathcal{Q}$, based on their similarity to $\mathbf{q}$:

\begin{equation}
    \mathcal{Q}^N = \text{TopK}(\{\text{sim}(\mathbf{q}, \mathbf{n}) \mid \mathbf{n} \in \mathcal{Q}\}, N),
\end{equation}

\noindent where $\text{sim}(\mathbf{a}, \mathbf{b}) := \mathbf{a}^\top \cdot \mathbf{b} / (\|\mathbf{a}\|_2 \cdot \|\mathbf{b}\|_2)$ is the cosine similarity and $\lVert \cdot \rVert_2$ denotes the $\ell_2$ norm. These hardest negatives serve as the basis for constructing synthetic contrasts that the model must learn to differentiate. 
We formulate synthetic hard negatives through a synthesis function, $\mathcal{F}: \mathcal{Z}\times\mathcal{Z} \rightarrow \mathcal{Z}$, that operates on the query and selected hard negatives:

\begin{equation}
   \mathbf{s}=\frac{\mathcal{F}(\mathbf{q}, \mathbf{n})}{\|\mathcal{F}(\mathbf{q}, \mathbf{n})\|_2}, \quad \mathbf{n}\in\mathcal{Q}^N,
\end{equation}

\noindent where $\mathbf{s}$ is the normalized synthetic negative. Following \cite{giakoumoglou2024synco}, we implement six different synthesis strategies for $\mathcal{F}$, including interpolation, extrapolation, mixing, noise jittering, perturbation, and adversarial methods.

The set of $L$ synthetic negatives, $\mathcal{Q}_s=\{\mathbf{s}_1, \mathbf{s}_2, \ldots, \mathbf{s}_L\}$, is then combined with the existing queue of real negatives, $\mathcal{Q}$, effectively expanding the diversity of negative examples. This combination of real and synthetic negatives allows the model to learn more robust and generalizable representations by providing more diverse and challenging contrasts during training, without requiring additional real data collection.









\section{Experiments} \label{sec:experiments}

We present comprehensive comparisons with state-of-the-art methods (Section~\ref{sec:main_results}) and subsequently conduct extensive experiments to evaluate our approach. We examine how synthetic data, $\mathcal{X}_s$, influences model performance (Section~\ref{sec:effect_synthetic_data}) and analyze the impact of synthetic hard negatives, $\mathcal{Q}_s$, on representation quality (Section~\ref{sec:effect_synthetic_negatives}).

\subsection{Setup}
\label{sec:setup}

We build on the implementations of MoBY \cite{xie2021moby} and SynCo \cite{giakoumoglou2024synco}. Specifically, we use MoBY as a baseline with the DeiT-Small \cite{touvron2021deit} and Swin-Tiny \cite{liu2021swin} architectures, following the same augmentation strategy as BYOL \cite{grill2020byol}. For synthetic negative generation, we implement all six strategies from SynCo \cite{giakoumoglou2024synco} and explore configurations with $N \in \{256,512,1024\}$ hardest negatives and synthetic ratios $L/(K+L)$. We conduct our experiments on ImageNet-100 \cite{deng2009imagenet,tian2020cmc}. For our synthetic data experiments, we generate a matching set of 130,000 images using the diffusion model from \cite{teng2024relay}, maintaining the same class distribution as the original dataset. We evaluate the quality of learned representations on ImageNet-100 classification using linear probing.

\subsection{Results} 
\label{sec:main_results}

\Cref{tab:main_results_table} presents a comprehensive evaluation of our proposed approach and comparisons with existing methods. Swin benefits primarily from synthetic negatives, achieving optimal performance with synthetic negatives alone. In contrast, DeiT effectively leverages both synthetic components, reaching peak performance with the complete approach during extended training. These architecture-specific responses highlight the importance of considering transformer design when applying synthetic augmentation strategies.

\subsection{Effect of Synthetic Data} 
\label{sec:effect_synthetic_data}

\Cref{fig:effect_synthetic_data} demonstrates that models learn effective representations from synthetic data alone, though performance improves with increasing real data proportions. Results indicate synthetic images serve as valuable complements rather than complete substitutes for real data. The relatively modest performance gap between fully synthetic and fully real training regimes suggests that modern diffusion models can generate high-quality samples that capture substantial semantic information from the original dataset.

\subsection{Effect of Synthetic Negatives} 
\label{sec:effect_synthetic_negatives}

\Cref{fig:effect_synthetic_negatives} shows architecture-specific responses to synthetic negatives. DeiT exhibits sensitivity to both hardness levels and proportions, while Swin maintains consistent performance gains across configurations.



\section{Discussion} 
\label{sec:discussion}

Our experiments show that architectural differences significantly impact how models leverage synthetic elements. The inclusion of synthetic data follows a diminishing returns pattern, which aligns with findings that diffusion models produce distribution shifts, limiting their effectiveness as complete substitutes. Synthetic negatives provide a computationally efficient enhancement, though their hyperparameters must be carefully tuned.


\section*{Acknowledgments}

We acknowledge the computational resources and support provided by the Imperial College Research Computing Service (\url{http://doi.org/10.14469/hpc/2232}), which enabled our experiments.


{
    \small
    \bibliographystyle{ieeenat_fullname}
    \bibliography{example_bibliography}
}

\end{document}